\def\year{2020}\relax
\documentclass[letterpaper]{article} 
\usepackage{aaai20}  
\usepackage{times}  
\usepackage{helvet} 
\usepackage{courier}  
\usepackage[hyphens]{url}  
\usepackage{todonotes}
\usepackage{graphicx} 
\usepackage{graphicx}  
\title{Hard Choices in Artificial Intelligence: \\  Addressing Normative Uncertainty through Sociotechnical Commitments}
\author{Roel Dobbe, Thomas Krendl Gilbert, Yonatan Mintz}
 
\begin{document}

\maketitle

\begin{abstract}
As AI systems become prevalent in high stakes domains such as surveillance and healthcare, researchers now examine how to design and implement them in a safe manner. However, the potential harms caused by systems to stakeholders in complex social contexts and how to address these remains unclear. In this paper, we explain the inherent normative uncertainty in debates about the safety of AI systems. We then address this as a problem of \textit{vagueness} by examining its place in the design, training, and deployment stages of AI system development. We adopt Ruth Chang's theory of \textit{intuitive comparability} to illustrate the dilemmas that manifest at each stage. We then discuss how stakeholders can navigate these dilemmas by incorporating distinct forms of \textit{dissent} into the development pipeline, drawing on Elizabeth Anderson's work on the epistemic powers of democratic institutions. We outline a framework of \textit{sociotechnical commitments} to formal, substantive and discursive challenges that address normative uncertainty across stakeholders, and propose the cultivation of related virtues by those responsible for development. 
\end{abstract}

\section{Introduction}


As the capabilities of AI expand, researchers must consider the problem of \emph{AI Safety}, defined broadly as the design of machines that can act independently while avoiding harm to people and the environment. This problem has become more urgent as these systems are technically refined, applied in critical infrastructure domains, and deployed in areas of social life not clearly related to physical harm (e.g. social media, credit scoring, public surveillance). Meanwhile, the core definitions of ``safety'' (and consequently appropriate control and governance mechanisms) in the wider literature of autonomous systems remain unsettled.


Distinct approaches to AI Safety have emerged to define the uncertain scale at which AI systems may cause significant social harm. At one end of this continuum is \emph{Existential Risk} (hereafter referred to as x-risk), i.e. the effort to mathematically formalize control strategies that help avoid the creation of systems whose deployment would result in irreparable harm to humans on a societal or civilization level. The x-risk literature has focused on the \emph{value alignment problem}; to ensure values programmed into an AI agent's reward function correspond with the values of relevant stakeholders (such as designers, users or others affected by the agent's actions) \cite{soares_value_2015}. Another approach, broadly pursued by researchers and increasingly social scientists in the \emph{Fairness, Accountability and Transparency in Computing Systems} (FAT*) literature, focuses on nearer-term problems broadly commensurate with existing social ills such as economic inequality, structural racism and gender disparities, and the capability of systems to (mis)recognize human affect or deny access to vital resources, among many other topics. FAT* research has harvested a multitude of definitions and tools aiming to address safety risks by \emph{diagnosing and reducing biases across various subgroups} defined along lines of race, gender or social class \cite{narayanan_fat*_2018}.

In both the FAT* and x-risk communities, a consensus is emerging on the fundamental limitations of technical approaches to formalize values such as safety or fairness in the design of AI systems, given the uncertainty of deployment contexts and the inevitability of externalities in using formal abstractions.
Within x-risk, \cite{hadfield-menell_incomplete_2018} have acknowledged the need for external social institutions to resolve ``unintentional and unavoidable misspecification'' in AI reward functions, and \cite{irving_ai_2019} propose to recruit social scientists to help resolve ``many uncertainties related to the psychology of human rationality, emotion, and biases''. Within FAT*, \cite{selbst_fairness_2019} pointed out the fundamental traps of abstraction that arise in formalizing notions of fairness in AI systems statistically and mathematically, while \cite{chouldechova_fair_2017} 
demonstrated 
tradeoffs arising in formalizing fairness that require moral deliberation. However, the criteria for evaluating and resolving harms remain vague across technical research communities, even as the call for an "algorithmic social contract" has crystallized \cite{rahwan2018society}.

Here we address the unavoidable challenge of \emph{normative uncertainty} in the development of AI systems, and articulate a set of commitments to address and resolve normative issues as they arise in any given development process.
We work from the assumption that AI systems are fundamentally of \emph{sociotechnical} nature, meaning they are built on and operated in contexts in which social and technical aspects are intimately interrelated.
This paper makes four contributions.
Firstly, we discuss normative uncertainty around the notion of safety by analyzing a case study, as conditioned by how different stakeholders interpret issues of protection, robustness, and resiliency.
Secondly, we outline existing responses to normative uncertainty in developing AI systems.
Thirdly, we introduce the philosophical concept of \textit{vagueness} to account for dominant intuitions about normative uncertainty in the AI Safety and FAT* literatures, as well as discussions thereof by critics and in the public sphere, as typifying distinct canonical approaches to dealing with vagueness (i.e. \emph{epistemicism, semantic indeterminism, and incomparability}). 
Lastly, to inspire the resolution of normative uncertainty in the development of AI systems, we draw on Ruth Chang's notion of \emph{intuitive comparability} (IC) to identify core dilemmas in designing, training and deploying AI systems.
We formulate a set of sociotechnical commitments, which address formal, substantive and discursive challenges, that are needed to consider values and stakeholders' needs in a democratic fashion.
In doing so, we formalize specific channels for dissent before, during, and after value commitments are being considered, building on the outline of democratic consensus-building in \cite{anderson_epistemology_2006}. 

Our core contribution is to apply an insight that scholars in Science \& Technology Studies (STS) have appreciated for over four decades: the reality that any technological system is inherently political and requires normative deliberation and ongoing citizen participation to ensure its safety for all stakeholders affected by its actions~\cite{winner_artifacts_1980}. 

\section{The Vagueness of Safety - ACLU vs. AWS}
\label{sec:the_vag_saf}
Safety has many definitions depending on context. 
For the purpose of study, we start with interpreting safety in terms of \textit{protection} from harm or injury, \textit{robustness} in the face of adverse conditions, and \textit{resiliency} in response to stress or difficulty. However, the vagueness of these terms as applied to different stakeholders makes it difficult for safety to be deliberated in a meaningful and consistent way. 

One example is the Amazon Web Services (AWS) Rekognition system. AWS intended Rekognition to be a commercially available cloud based ML tool that helps enterprises with setting up and searching image based datasets for various facial recognition tasks \cite{amazon_rekognition_2019}. By design, the system is flexible and allows users to define their own data sets and queries, since it is meant to perform well for tasks as simple as document retrieval and as complex as image-based sentiment analysis. To test the system's limitations, the American Civil Liberties Union (ACLU) created a data set made up of publicly available arrest photos and used these to train Rekognition. They then queried the system to find a match against photos of current members of Congress \cite{snow_amazons_2018}. The ACLU reported not only that false identifications were found, but that of those members of Congress falsely identified as matching the arrest photo database, a disproportionate number (40\%) were people of color. The ACLU concluded that Rekognition shows a bias in its predictions, making it unsafe to implement in high stakes contexts which disproportionately affect people of color (e.g. law enforcement). However, the AWS research team issued a rebuttal, commenting that Rekognition was used against its articulated design specifications \cite{wood_aws_2018}. In particular, they noted that the ACLU used a lower-than-recommended confidence threshold for a high risk task, leading to a large number of false positives. Moreover, since the ACLU did not detail how they constructed their data set, it is unclear how much inherent bias existed in the queried set. 
Below we detail the forms of vagueness present in this scenario and identify relevant trade-offs.

First, it is unclear what \textit{protection} means in development situations where private and public definitions are both at stake. Concretely, many AI systems are validated internally by private corporations, only to be misused in deployment due to poor communication about the inner workings of the system to the wider public. In part, the conflict between Amazon and the ACLU is based on a category mistake in how the boundaries of different political guarantees are refashioned by the development norms of the system in question. For example, the ACLU's claim that facial recognition systems beneath a given accuracy threshold should not be used by law enforcement is rooted in the intuition that correcting misclassifications \textit{a posteriori} is politically unacceptable, as it imposes Amazon's internal definitions of \textit{harm} and \textit{vulnerability} onto anyone that encounters the system. In contrast, Amazon's claim that the system does work according to design intentions and that the ACLU study used inappropriate settings makes sense in the context of system optimization, as the AWS team converged on the system's architectural parameters through agreements with the private contractors who intend to use it. Fundamentally, it is not clear whether the safety of Rekognition should be determined by its protection of \textit{private contracts} (whose context is the online verification of edge cases by self-interested parties) or \textit{public assurances} (whose context is the willingness to shield vulnerable communities from harm). While both Amazon and the ACLU value protection, the legal contexts for their practical definitions of it are orthogonal, and the loci of perceived stakes are at two different points in the development pipeline. To resolve this vagueness, either one definition must be given absolute priority over the other, or formal distinctions across the pipeline must be clarified and subject to a consistent, external standard.


Second, conditions of \textit{robustness} must be specified according to the distinct expectations of designers and users, leading to inconsistent standards for platform governance. One can imagine AWS issuing a different response that included an apology to members of Congress, a request for the ACLU to expand its "testing" to other social domains, and a promise to improve Rekognition's accuracy going forward. However, this strategy might also become an object of public outcry; for example, Waymo regularly publishes safety reports on its vehicles but still faces the ire of Phoenix residents, who complain that "They didn't ask us if we wanted to be part of their beta test" \cite{romero_wielding_2019}. Indeed, a handful of American cities have now banned the use of facial recognition by municipal agencies, citing surveillance concerns, local community interests, and social prejudice \cite{lee_sanfran_2019}. The question is whether cities and other social domains should be made ready for facial recognition tools (through e.g. concrete institutional reforms), or the tools should be made ready for cities to use them with confidence (via e.g. ongoing technical refinement). How one answers this question places the onus of sociotechnical robustness on either the \textit{public officials} who administer the system or the \textit{engineers} who build it, a problem known as defining the "moral crumple zone" of moral and legal responsibility \cite{elish_crumple_2019}. 
While all stakeholders might want the system to be robust, it is not clear what criteria should determine the conditions under which robustness would hold, and which authorities are most qualified to ensure those conditions. Because system development transposes inherited notions of governance and control, specific liability mechanisms (e.g. moratoriums, audits) cannot be adopted or justified until these notions are clarified and made compatible.

Third, a system's \textit{resiliency} requires a metric of optimality, according to which abnormal dynamics can be discerned, diagnosed, and remedied. At a minimum, facial recognition assumes some definition of what a face \textit{is}, and what the good, bad, and inaccurate ways there may be for identifying them. While the dispute between AWS and the ACLU did not reach this level of abstraction, we must confront how AI systems may affect our political sovereignty and rewire social orders by shifting how human features are modeled, correlated, and interpreted. Recently, Luke Stark has argued that facial recognition tools are a form of racism that will incline any power structure towards discriminatory policies because the ability to rank facial features at scale will generate categories that can be used both to solder somatic attributes to personality characteristics and to legitimize political decisions \cite{stark2019}. To avoid this future, such tools should be regulated to the point that they are hardly ever used. Meanwhile, \cite{wang2017} claim to detect sexual orientation with the aid of deep neural networks and that the predictive power of AI models can be harnessed to discover patterns in facial features beneath human awareness. While the findings generated controversy \cite{murphy2017,vincent2017}, Kosinski defended the study as an effort to ``understand people, social processes, and behavior better through the lens of digital footprints'' \cite{resnick2018}, questioning whether automated systems can only reify existing social ontologies or substantively challenge current intuitions about gender and sexuality. This contrast highlights the vague relationship between feature orderings in particular contexts, and the general goals or ends that define human flourishing. The safety of a facial recognition system is determined by how its dynamics are defined in light of overarching risks and benefits: as \textit{illegitimate} (making resiliency impossible) or \textit{legitimate} (defining a sovereign metric for acceptable uses and outcomes). The deliberation behind this amounts to what kind of society is wanted.

\section{Resolving Vagueness via Hard Choices}
\label{sec:vagueness}
We draw from Ruth Chang's philosophical work on value pluralism~\cite{chang_incommensurability_1997,chang_possibility_2002} to develop a sociotechnical semantics for AI Safety. At certain deployment scales, what we mean by ``privacy'', ``security'', or ``social choice'' starts to feel unclear, as it is difficult to determine the stakes of our own value commitments. In philosophical terms, the relations between our values become \textit{vague}. 

Chang outlines three distinct approaches to vagueness: (1) \emph{epistemicism} - all items of value can be ranked against each other in order for vagueness to be resolved, (2) \emph{semantic indeterminism} - the way values relate to each other is fundamentally fuzzy, and (3) \emph{value pluralism} - values are incomparable. In the Appendix, we further introduce these approaches and tie them to recent work in addressing normative uncertainty in the development of AI systems.

Here we introduce an alternative perspective, which forms the basis for our proposal of sociotechnical commitments in the AI development process. \cite{chang_hard_2017} proposes a fourth position, \textit{intuitive comparability} (IC): while many human values seem incommensurable (e.g. equality vs. liberty), humans are nevertheless able to articulate \emph{evaluative differences} to make comparisons, even if two values or concepts are not directly measurable against each other. This allows people to make informed tradeoffs between options (e.g. choosing between a banana and a donut for breakfast) based on practical deliberation regarding one's overarching goal (losing weight on a diet), even though \textit{nutrition} and \textit{tastiness} are qualitatively distinct--the values are objectively incommensurable but, in this context, intuitively comparable. IC is particularly relevant for what she calls \emph{hard choices}: when different alternatives are \emph{on a par}, ``it may matter very much which you choose, but one alternative isn't better than the other [...] alternatives are in the same neighborhood of value, in the same league of value, while at the same time being very different in kind of value \cite{chang_hard_2017}. 
Resolving hard choices requires normative reasoning: ``when your given reasons are on a par, you have the normative power to \emph{create} new will-based reasons for one option over another by putting your agency behind some feature of one of the options. By putting your will behind a feature of an option - by standing for it - \emph{you} can be that in virtue of which something is a will-based reason for choosing that option.''

\subsection{The Case for Intuitive Comparability in AI Safety}
We endorse IC and parity not as a superior philosophical position in opposition to others, but as a lens from which to ask and analytically identify what a cohesive approach to AI Safety would look like. In the development of AI systems, IC provides a foundation for normative reasoning between possible value regimes. Hard choices cannot be determined through purely quantitative thresholds; instead, IC suggests the iterative redrawing of a system's formal boundaries and design parameters via qualitative feedback. The ``hard choice'' moments are those where different communities, comprising the affected stakeholders, may clash and new development criteria (what Chang calls ``will-based reasons'') must be specified. This matches the intuition that the design of AI systems restructures the context in which users or other affected stakeholders exist: ``values emerge, whether you look for them or not''~\cite{halloran_value_2009}. 
\cite{iversen_rekindling_2010} argue this requires an ``\textit{a priori} commitment to cultivate the emergence and discovery of local expressions of values whilst being mindful of further expression of values during the course of the design process''.
In addition, the value hierarchy designed into a system will determine the space of actions available to it (as well as those that the system forecloses), and it is crucial to acknowledge and account for the power and elevated status of design work.
This means recognizing developers' tendencies to prioritize certain actors and networks over others. 
\cite{haraway_situated_1988}, \cite{harding_science_1986} and other feminist scholars would argue that we cannot escape having some agenda: after all, the researcher is also situated in the social world they study. 
A crucial corollary of the above is that developers have the responsibility to take a political stance. As Ben Green notes, ``to remain apolitical is itself a political stance - a fundamentally conservative one (in the sense of maintaining the status quo rather than in relation to any specific political party or movement) - and why the field's current attempts to promote `social good' dangerously rely on vague and unarticulated political assumptions''~\cite{green_data_2018}.

AI Safety also presents two sources of nuance to hard choices. Firstly, AI systems need to encode hard choices made by or on behalf of a diverse group of stakeholders affected by the system,
including divergent values and interests. These are fundamentally political, which is well understood in the STS literature~\cite{winner_artifacts_1980}. Our goal is to build an analytical framework to draw attention to these moments and facilitate bridges between the ongoing contributions of AI Safety research and the core substantive insights of STS scholarship. Secondly, the values that stakeholders care about are often complex and not readily translated into a solution that suits all needs, which can lead to situations of \emph{moral overload} that require thinking outside of the traditional design space~\cite{van_den_hoven_engineering_2012}. As such, we extend and elaborate the argument of \cite{hadfield-menell_incomplete_2018} that acknowledges the need to address misspecification between reward functions and wider social institutions.

Those responsible for developing and governing AI systems have the duty to mediate hard choices and the corresponding value conflicts \emph{across} different stakeholders, allowing them to resolve these choices through both quantitative and qualitative evaluation. 
Like the maintenance of cables that span a suspension bridge, AI Safety can be defined as the successful maintenance of the relations that comprise the conceptual space of comparability for human values across all stakeholders. Just as the ``stress point'' of civil engineering is the identified and agreed-upon maximum strain the bridge can handle before buckling, the critical point for human-compatible AI is the safeguarding of comparability, i.e. the capability of AI systems to support pluralist value hierarchies while preserving \emph{shared moral agency}; the power to engage in design, training, and deployment.




\section{Sociotechnical Commitments in Developing AI}

We propose a set of commitments that situate the design, training and deployment stages of AI systems in their sociotechnical context and center and address issues of vagueness. The commitments are comprised of formal, substantive and discursive challenges to the development process in order to safeguard stakeholders' access to hard choices in system development.
Formally, these challenges comprise distinct tradeoffs that are unavoidable in the agonistic development of systems that must inherit, translate, and instantiate conflicting values. Substantively, they condition our value commitments in a manner that is not zero-sum, extending the boundaries of the system and design space to ensure the expression of intuitively comparable human values and resolve situations of moral overload~\cite{van_den_hoven_engineering_2012}. 
Discursively, they compel communication between stakeholders: formulating the problem, evaluating systems that would solve it, and articulating the values and needs that the system must address in order to be safe. We posit that such ongoing stakeholder engagement requires ``reflexive inquiry [that] places all of its concepts and methods at risk [...] not as a threat to rationality but as a promise of a better way of doing things''~\cite{agre_toward_1997}.

Following \cite{anderson_epistemology_2006}, we emphasize the need for dissent mechanisms during the design, testing, and implementation of automated systems. Tracking dissent is necessary in order to respect the IC of available value hierarchies while reconceiving AI development as an opportunity to reimagine the moral communities to which we belong. This choice is in itself normative, and may inspire particular legal translations depending on the application domain and the jurisdiction and democratic regime in which a system is built. 
In many instances, regulatory measures may form either an existing source of constraints and requirements in the development process, or be informed by it. 
The authors do not advocate particular law or policy interpretations, but see such translation work as a natural extension of this paper. 

To illustrate our sociotechnical commitments, we will refer to the AWS-ACLU case study described earlier. We consider the relevant hard choices made throughout the design, training, and deployment of the Rekognition system and illustrate their political impact. At each stage, we ask: (1) how does vagueness arise and what forms may IC take? (2) In what concrete ways can \textit{formal affordances}, \textit{substantive commitments}, and \textit{discursive practices} address these issues?

\subsection{Design}
AI systems generally represent a predictive model that can be trained and used in the decision making capabilities of some human agent or automated control system.
As the model represents an \emph{abstraction} of the phenomenon about which it makes predictions, the chosen model parameterization and the training data used to determine parameter values delimit the possible value hierarchies that may be encoded and, if not anticipated and accounted for, may deny stakeholders the opportunity to evaluate design alternatives and force potentially harmful and unsafe hard choices.
To harness IC in the design stage, the following challenges must be taken up: (1) \textit{Formal challenge}: Make explicit and negotiate what can and cannot be modeled and inferred, crystallized in the model-based/model-free dilemma;
(2) \textit{Substantive challenge}: Make a modeling commitment whose application constraints leave flexibility for different stakeholders to forge their own values during training and deployment;
(3) \textit{Discursive commitment}: validate the design with stakeholders to anticipate possible value conflicts that can arise due to the gap between model and world and plurality of values during deployment, preparing for design iterations.


The design stage determines the computational powers of the system: how the limits of what it can model determine its assumptions about people and what kinds of objects or classes (e.g. faces) are recognizable to it. 
At a minimum, stakeholders must answer the following: 

\textbf{Model-based:} What domain knowledge is available to model the environment? I.e. what is the permissible space in which a given problem can be formulated and solved, and what modeling tools are available? 

\textbf{Model-free:} What are permissible predictive signals within the environment? I.e. what are the base rewards, elements, or qualities that could shape the system's policies? How should these take qualitative precedence over others? 

Formally, the dilemma manifests in choosing a model capacious enough to represent the nature of the environment, but constrained enough that its training would not be intractable.
Imposing modeling constraints also creates \textit{technical bias}, which may take away space for stakeholders to express or protect their own specific values in terms of the phenomena permitted or excluded by the model's system boundaries. In the context of our case study, the dilemma rests in the choice between giving Rekognition some hard-coded limitations on how the algorithm may be used, vs. permitting the algorithm to extract whatever signals it needs in order to maximize its predictive accuracy. While AWS did not address this in its response to the ACLU, they could have done so accordingly: either transition to a more automated decision procedure that sacrifices direct human oversight but is more accurate for the congressional dataset at stake, or propose a governance structure to mediate the ethical and legal applicability of the tool and ratify the environmental conditions a system is allowed to represent (here, the features appropriate for recognizing representatives' faces).

Either way, the dilemma is resolved via \textit{context discernment}, the disqualification of specific features and actions within the problem space in advance of deployment. 
Here we draw from
\cite{dreyfus_all_2011}: ``The task of the craftsman is not to \textit{generate} the meaning, but rather to \textit{cultivate} in himself the skill for \textit{discerning} the meanings that are \textit{already there}''.
Design teams need to consider how the algorithm is expected to be integrated in and interacting with the context of deployment, what bias issues may arise during training and how to account for and protect vulnerable user groups, and how chosen objective functions may generate externalities, as well as who is likely to bear their cost.
In the event no consensus is reached and dissent persists, the option of not designing the system should remain viable.

\subsection{Training}
After certain features and ways of modeling have been disqualified through design commitments, the specific weights and the structure of the predictive mapping must be determined by performing an optimization problem. This determines the input-output behavior of the model and how it will interact with human agents and other systems.
Through the recruitment of historical and experimental data, the system can (1) infer causal model parameters, (2) infer parameters of noncausal representations, and (3) iteratively adjust parameters based on ongoing experiments (as in reinforcement learning).
To harness IC in the training stage, the following challenges have to be taken up:
(1) \textit{Formal challenge}: Assess the limits of how parameters can be inferred at present, crystallized in the validation/verification tradeoff;
(2) \textit{Substantive challenge}: Make a validation commitment that is acceptable to present stakeholders;
(3) \textit{Discursive challenge}: Form a team consensus around verification strategies to be pursued during deployment and define alternate design strategies that might aid parameter inference.


The system must be trained to bridge the gap between features by generating correlates: it must take in data, update priors, and handle edge cases. This is done with the help of engineers who interface between the problem the system is meant to solve and the workings of the system itself. 
Here, the minimum requirements for certifying safe outcomes are \emph{impartial assessments} of the following questions:

\textbf{Verification:} Was the right system built? Are the needs of prospective users being met? Is the specified problem solved or not? Is the system able to predict what it was meant to?

\textbf{Validation:} Was the system built right? Are there hidden utility monsters or emergent biases? Is there risk of strategic behavior or manipulation? What information channels must be provided to users to minimize these likelihoods?

%
Systems whose models are made more accurate or robust for well-specified environments (and subpopulations of people) will be made brittle and possibly unworkable for poorly-defined environments, which can result in disparate impacts, especially among yet underrepresented (and undersampled) minorities that already face systemic marginalization and are not properly represented on AI design teams~\cite{west_discriminating_2019}. This effect can clearly be seen in the case between AWS and the ACLU. Here, the designers of Rekognition created the system with common commercial tasks in mind (e.g. sentiment analysis), and determined their own confidence levels through extensive internal verification. However, the ACLU deployed Rekognition to a task that it was not explicitly meant to perform, i.e. matching faces of politicians to those of recent arrestees, and indicated a weakness in the way the system was validated. 

A commitment to \textit{team consensus} is needed to integrate the problems of value alignment into standard procedures for quality assurance. This is achieved by forcing agreement among system engineers about how to allocate sparse team resources between system verification and validation in order to manage under-specification risks and mitigate the perversion of intended users' semantic and moral commitments. The team must decide: what commitments to contracted users are necessary for the desired balance of model testing to be adequate? Here ``quality management'' must be elevated to the contestation and adjudication of how (possibly pluralist) values are operationalized without compromising comparability. In the case of Rekognition, AWS implemented a confidence threshold slider with additional documentation commenting on how it should be set for different contexts that will require distinct value hierarchies (e.g. law enforcement vs. automatic image tagging) \cite{amazon_rekognition_2019}. However, per the ACLU comment in response to Amazon's defense of the system \cite{aclu_2018}, it is unclear whether this specific metric is sufficient to validate consensus among all relevant stakeholders (e.g. members of Congress, government agencies, Amazon employees) rather than necessary for system verification alone, as some of its confidence thresholds appear arbitrary.

\subsection{Deployment}
Finally, the system must define use cases in terms of a user contract that identifies terms of consent and ensures interpretive understanding without coercion. The resulting deployment conditions determine the authority of the system's representations in the context of user agency, i.e. what the user wants the system to be for them.
Here we appropriate tradeoffs already identified by social theorists regarding the moral authority and political powers of social institutions~\cite{flew_citizens_2009}.
To preserve IC in the deployment stage, the following have to be taken up: (1) \textit{Formal challenge}: Assess what kind(s) of agency users have if the verification fails, crystallized in the exit/voice dilemma; (2) \textit{Substantive challenge:} Make a commitment to an open feedback channel by which users express their values on their terms; (3) \textit{Discursive challenge:} Justify that channel by means of a public commitment to users that establishes that channel as trustworthy.

Resolving these challenges requires \emph{representative input and mitigation of issues} for the following:

\textbf{Exit}: Are users able to withdraw fully from using the product or platform? Is there any risk in this? Are there competing products or platforms they can use? Have assurances been given about user data, optimization, and certification after the user withdraws?

\textbf{Voice}: Can users articulate proposals in a way that makes certain concerns a matter of public interest? Are clear proposal channels provided for users, and are they given the opportunity to contribute regularly? Are the proposals highlighted frequently considered and tested, e.g. through system safety? Are users kept informed and regularly updated?

To the extent that proposed value hierarchies remain indeterminate after the commitments made during design and training, deployment challenges systems to handle the multiple objectives, values, and priorities of diverse users. At stake here are the unexpressed moral commitments of subpopulations not originally considered part of the potential userbase, who must bear the ``cost function'' of specification. 
Concretely, deployment administrators must determine whether the user will interpret the system agreement as primarily economic (in which case the user acts as a \textit{consumer}) or political (in which case the user acts as a \textit{citizen}). More Exit implies a market setting, while more Voice suggests a political context. For example, the user agreements of Rekognition may be understood either in terms of \textit{private contracts} (in which case data is treated as a commodity and alternative platforms are implied to exist) or \textit{public assurances} (in which case data is inalienable and Rekognition is interpreted as a public utility or service). 
If the former takes precedence, Rekognition's deployment might be regulated with \textit{private certification} that attests the features and uses of data it forbids; if the latter is more important, deployment increasingly depends on a \textit{public accreditation} that guarantees the user's legal protections will take priority in all use cases regardless of features or data.

Deployment administrators and their regulating authorities must cultivate \textit{public accountability} to deal with these challenges, ensuring both Voice and Exit remain possible for users such that some form of accountability is maintained: anyone can leave if they want, but enough people choose to remain because they trust in their ability to express concerns as needed. This balance must hold regardless of the specific commitment being made--for example, AWS may specify some channel by which vulnerable groups can opt out of a publicly-operated Rekognition use context (preserving Exit), or supply private contractors with a default user agreement that must be relayed to anyone whose data will be used by the system (preserving Voice).
Either way, administrators should model model people neither as \textit{consumers} (a customer, client, or operator treated more or less as a black box) nor as \textit{citizens} (a subject with guaranteed rights, among them the right to dissent to relevant forms of political power) without making the commitment explicit as justification for the terms of deployment.

\section{Conclusion}
Clarifying the sociotechnical foundations of safety requirements for AI systems will lay the groundwork for system developers to take part in distinct dissent channels proactively, before the risks posed by AI systems become technically or politically insurmountable. 
We anticipate this set of sociotechnical commitments will need to be integrated into the training of engineers, data scientists, and designers as qualifications for the operation and management of advanced AI systems in the wild. Ultimately, the public itself must be educated about the assumptions, abilities, and limitations of these systems so that informed dissent be made desirable and attainable as systems are being deployed--deliberation is the \textit{goal} of AI Safety, not the procedure by which it is ensured.
We endorse this approach due to the computationally underdetermined, semantically indeterminate, and politically obfuscated value hierarchies that will continue to define diverse social orders both now and in the future. Democratic dissent, as a pathway to system development, is necessary for such systems to safeguard the possibility of IC and allow users to define the contours of their own values. To paraphrase Reinhold Niebuhr, AI's capacity for value alignment makes development commitments possible, but its inclination to misalignment makes commitments necessary.

\bibliographystyle{aaai}
\bibliography{references_aisafety}

\appendix

\section{Prominent Approaches to Vagueness}
Below we briefly present four prominent approaches to vagueness as attempted responses to the problem of normative uncertainty, and tie them to existing vital approaches in the AI Safety literature.

In \textit{Epistemicism}, all items of value can (and must) be ranked against each other in order for vagueness to be resolved; even if some things are fundamentally ``apples and oranges'', there must be some degree to which one is preferable over the other, however much regret or compromise must be faced as a consequence. Within the x-risk literature, a recent epistemicist approach to resolving normative uncertainty is \textit{metanormativism}. As proposed by William MacAskill \cite{macaskill_normative_2014}, metanormativism seeks to articulate "second-order norms" that guide how one should act when multiple appealing moral doctrines are available--for example, whether or how an advanced AI system capable of transforming individual/group/civilization utility should be designed. As one example, MacAskill, whose work has been cited in support of technical work on value alignment
and value learning~\cite{soares_value_2015}, has proposed a ``choice worthiness function'' that would generate reward functions in an ``appropriate'' manner, where \textit{appropriateness} is defined as ``the degree to which the decision-maker ought to choose that option, in the sense of ‘ought’ that is relevant to decision-making under normative uncertainty''~\cite{macaskill_normative_2016}. In a safety context, such a function would provide guidance for AI theorists and designers about what decision procedures are acceptable or unacceptable for the system to follow, specifically when the goal state and risk scale are difficult to define.

Metanormativism still requires the existence of a clear, positive value relation between available ethical actions: one must be unambiguously better, worse, or equal to the other for the choice worthiness function to hold. In this way, Epistemicism holds out for some ``Platonic value function'' that defines an absolute hierarchy of values which, while currently unknown, can be solicited and learned by sufficiently-capable automated systems.
One example in the context of AI Safety is recent work by \cite{hadfield-menell_incomplete_2018} on incomplete contracting, in which suggestions are made to let robots replicate the external \emph{normative social order} to overcome unavoidable misspecification of human objectives in AI reward functions. 
Similarly, \cite{irving_ai_2019} propose addressing misspecification by ``asking people questions about what they want, training machine learning (ML) models on this data, and optimizing AI systems to do well according to these learned models'', rather than looking beyond the confines of the reward function.

But there is a wider problem of defining what safety means throughout the development pipeline that transcends the epistemic uncertainty about which ethical norms should guide design decisions. The examples in the section on The Vagueness of Safety show ``safety'' issues, particularly once AI systems expand in their application, cannot be resolved through metanormative guidance at the design stage alone, and require a wider sociotechnical diagnosis of the entire development pipeline. Still, what the normative foundations of AI Safety \textit{should be} remains unclear. On one hand, most technical AI Safety aims to capture humans values through a single, partially-specified reward function, either \textit{a priori} or through ongoing human-agent interactions. Yet even carefully-designed formalisms that are sensitive to the implicit concerns and priorities of human agents are not guaranteed to learn the right preference structures in the right way without new forms of surveillance, control, and assigned roles for both humans and the systems themselves, see for instance~\cite{hadfield-menell_cooperative_2016}.
Consider also the proposal by \cite{eckersley_impossibility_2018} that, given the impossibility of simultaneously maximizing total wellbeing and average wellbeing while also avoiding suffering, ``the human is
\textit{torn} between objectives that fundamentally cannot be traded
off against each other''. Such system setups are limited in three key ways: (1) they can never formalize everything, and require subsequent developers to organize around them; (2) they attempt to resolve both content and procedure from the get-go, rather than treat the sociotechnical development of AI systems as a dynamic and iterative problem; (3) they are limited in addressing the wider spectrum of values across distinct peoples and cultures.


\emph{Semantic indeterminacy}, however, asserts a fundamental ambiguity or arbitrariness in how values are related to each other. This position is practically adopted by software engineers and tech enthusiasts for whom value indeterminacies comprise an investment opportunity for new AI systems. The following quote from Mark Zuckerberg is illustrative: ``I'm also curious about whether there is a fundamental mathematical law underlying human social relationships that governs the balance of who and what we all care about [...] I bet there is'' (quoted after~\cite{hildebrandt_privacy_2019}).

The danger here is that software engineers may arbitrarily neglect certain value hierarchies, either by failing to interrogate the context of historical data or external cost biases through design choices that moralize existing structural inequalities ~\cite{eubanks_automating_2018}. As such, this position neglects the \textit{public interest} as potentially compromised by the external costs of society-wide technological innovations, a problem at the heart of democratic political theory~\cite{dewey_public_1954}.


\textit{Value pluralism} holds that there cannot or will never be an ultimate scheme for organizing values against each other, because our reasons for holding them are incomparable. Notably, \cite{macaskill_infectiousness_2013} defines ``infectious'' incomparability as equivalent to nihilism, given its ``view that the notions of good and bad and of right and wrong are illusions and that, objectively speaking, no option or state of affairs is better than any other, nor are any two options or states of affairs equally good''. However, value pluralism is widely adopted by critical theorists who highlight how formal value specifications, no matter how well-intentioned, always end up excluding certain subpopulations in favor of others. Furthermore, in \emph{any} system design, fundamental choices have to be made about how values of relevant stakeholders, including those indirectly affected by the system, result in some \emph{value hierarchy} that has real consequences for how the benefits and harms of a system play out across society. \cite{crawford_can_2016} endorses the concept of \emph{agonistic pluralism} as offered by \cite{mouffe_deliberative_1999} as a design ideal for engineers: ``[T]heories of agonism allow us to widen the perspective to include the contested spaces where algorithms are designed. They are always made by and in relation to people: they are in flux and embedded in hybrid spaces. [...] These workplaces are themselves spaces of everyday conflict and dissent, where algorithmic design decisions are made after debate, disagreement, tests and failures.'' Meanwhile, \cite{hoffmann_where_2019} argues that fairness approaches fail to address the hierarchical logic that produces advantaged and disadvantaged subjects: ``by centering disadvantage, we fail to question the normative conditions that produce - and promote the qualities or interests of - advantaged subjects,'' and disproportionately put safety harms on already vulnerable populations. \cite{hildebrandt_privacy_2019} takes these perspectives to their logical extreme and advocates for \textit{agonistic machine learning}, suggesting that the human self should be treated as fundamentally incomputible.

Moreover, value pluralism is not opposed to external mechanisms that regulate how our diverse commitments may be reconciled~\cite{james_will_1979}.

\textit{Intuitive Comparability} \cite{chang_hard_2017} instead endorses a fourth position called \textit{intuitive comparability} (IC): while many human values seem incommensurable (e.g. equality vs. liberty, good nutrition vs. gourmandism), humans are nevertheless able to articulate \emph{evaluative differences} to make comparisons, even if two values or concepts are not directly measurable against each other. IC is particularly relevant for what she calls \emph{hard choices}: when different alternatives are \emph{on a par}, ``it may matter very much which you choose, but one alternative isn’t better than the other [...] alternatives are in the same neighborhood of value, in the same league of value, while at the same time being very different in kind of value \cite{chang_hard_2017}. Resolving hard choices requires normative reasoning: ``when your given reasons are on a par, you have the normative power to \emph{create} new will-based reasons for one option over another by putting your agency behind some feature of one of the options. By putting your will behind a feature of an option—by standing for it—\emph{you} can be that in virtue of which something is a will-based reason for choosing that option.''~\cite{chang_hard_2017}

We propose IC as an intermediary position that serves as a stepping stone to a complete sociotechnical depiction of normative uncertainty, one that will need to combine the intuitions of technical AI Safety and critical perspectives
It can accommodate the pluralist perspectives and value commitments of marginal stakeholders by means of development standards that help reconcile political power with technical innovation, all while avoiding defeatism. 
It entails neither a single nor infinite number of reward functions, but rather a capacity for mediating different definitions of safety in the context of systems that may know our values differently than we know ourselves. Consequently, abstract criteria for \textit{appropriateness} may not be applicable to AI Safety until the philosophical problem of \textit{vagueness} is addressed throughout the development pipeline by incorporating the hard-learned lessons of \emph{values and ethics in human-computer interaction design} (see~\cite{shilton_values_2018} for an extensive overview) and STS, 
which the AI Safety communities have yet to absorb.

These conclusions also find support in the field of \emph{Computer Supported Cooperative Work} (CSCW). Presented as the field's central intellectual challenge, \cite{ackerman_intellectual_2000}
described the inevitability of the \emph{social-technical gap} of computer systems; the inherent divide between what we know we \emph{must} support socially and what we \emph{can} support technically.

\end{document}